\documentclass[conference]{IEEEtran}
\IEEEoverridecommandlockouts

\usepackage{cite}
\usepackage{amsmath,amssymb,amsfonts}
\usepackage{algorithmic}
\usepackage{graphicx}
\usepackage{textcomp}
\usepackage{xcolor}
\usepackage[T1,T2A]{fontenc}
\usepackage[utf8]{inputenc}
\usepackage[english]{babel}
\usepackage{url}

\usepackage{caption}

\usepackage{tabularx} 
\usepackage{booktabs}

\def\BibTeX{{\rm B\kern-.05em{\sc i\kern-.025em b}\kern-.08em
    T\kern-.1667em\lower.7ex\hbox{E}\kern-.125emX}}
\begin{document}

\bibliographystyle{IEEEtran}

\title{Building a Macedonian Recipe Dataset: Collection, Parsing, and Comparative Analysis}

\author{\IEEEauthorblockN{1\textsuperscript{st} Darko Sasanski}
\IEEEauthorblockA{\textit{Faculty of Computer Science and Engineering} \\
\textit{Ss. Cyril and Methodius University}\\
Skopje, North Macedonia \\
darko.sasanski@finki.ukim.mk}
\and
\IEEEauthorblockN{2\textsuperscript{nd} Dimitar Peshevski}
\IEEEauthorblockA{\textit{Faculty of Computer Science and Engineering} \\
\textit{Ss. Cyril and Methodius University}\\
Skopje, North Macedonia \\
dimitar.peshevski@finki.ukim.mk}
\and
\IEEEauthorblockN{3\textsuperscript{rd} Riste Stojanov}
\IEEEauthorblockA{\textit{Faculty of Computer Science and Engineering} \\
\textit{Ss. Cyril and Methodius University}\\
Skopje, North Macedonia \\
riste.stojanov@finki.ukim.mk}
\and
\IEEEauthorblockN{4\textsuperscript{th} Dimitar Trajanov}
\IEEEauthorblockA{\textit{Faculty of Computer Science and Engineering} \\
\textit{Ss. Cyril and Methodius University}\\
Skopje, North Macedonia \\
dimitar.trajanov@finki.ukim.mk}
}

\maketitle

\begin{abstract}
Computational gastronomy increasingly relies on diverse, high-quality recipe datasets to capture regional culinary traditions.  Although there are large-scale collections for major languages, Macedonian recipes remain under-represented in digital research. In this work, we present the first systematic effort to construct a Macedonian recipe dataset through web scraping and structured parsing. We address challenges in processing heterogeneous ingredient descriptions, including unit, quantity, and descriptor normalization. An exploratory analysis of ingredient frequency and co-occurrence patterns, using measures such as Pointwise Mutual Information and Lift score, highlights distinctive ingredient combinations that characterize Macedonian cuisine. The resulting dataset contributes a new resource for studying food culture in underrepresented languages and offers insights into the unique patterns of Macedonian culinary tradition.
\end{abstract}


\section{Introduction}

Food is a central element of cultural identity, and researchers are increasingly applying computational methods to the study of culinary traditions in the emerging field of computational gastronomy~\cite{min2019survey}. Recipe datasets provide a valuable resource for applications ranging from ingredient substitution and recipe recommendation to nutritional analysis and cross-cultural studies of cuisine~\cite{sasanski2025fooddatasemanticweb, trattner2017food}. The quality, scale, and linguistic diversity of available datasets critically influence the scope of such research.

In recent years, large-scale English-language resources have formed the foundation for much of this work. In particular, Recipe1M+~\cite{marin2021recipe1m+} introduced more than one million recipes paired with images, enabling advances in cross-modal learning and food recognition. Building on such resources, FoodKG~\cite{haussmann2019foodkg} further enriched recipe data by aligning ingredients and instructions with structured ontologies, demonstrating the potential of semantic representations for tasks such as knowledge graph construction and reasoning over food data. These datasets have been widely used to study culinary patterns~\cite{ahn2011flavor}, generate recipes~\cite{salvador2019inverse}, and explore links between food, health, and culture~\cite{min2019survey,trattner2017food}.

Despite these successes, existing resources are heavily biased toward major languages and international cuisines~\cite{marin2021recipe1m+}. Regional and underrepresented culinary traditions remain largely absent from computational research, limiting both the inclusivity of global food analysis and the development of tools tailored to local contexts. Macedonian cuisine, with its rich history and distinctive combinations of ingredients and cooking techniques, exemplifies such an underrepresented tradition. To date, no publicly available dataset captures Macedonian recipes in a structured, machine-readable form.

This work presents the first systematic effort to construct a comprehensive Macedonian recipe dataset. Recipes were collected via web scraping and processed through structured parsing to normalize heterogeneous ingredient descriptions. Key challenges include normalizing variations in units, quantities, and descriptive modifiers. Beyond dataset construction, we perform an exploratory analysis of ingredient usage and co-occurrence patterns. Using statistical association measures such as Pointwise Mutual Information (PMI)~\cite{church1990word} and Lift score~\cite{agrawal1993mining}, we highlight distinctive ingredient pairings and combinations that characterize Macedonian cuisine.

Our contributions are threefold. First, we introduce the first structured dataset of Macedonian recipes. Second, we describe the challenges and solutions in parsing ingredient descriptions for an underrepresented language. Third, we provide an exploratory analysis that reveals unique features of Macedonian culinary tradition.

This dataset and analysis extend the coverage of computational gastronomy to a new linguistic and cultural domain, supporting future research on food culture, nutrition, and regional cuisines.

\section{Dataset Construction}

To enable computational analysis of Macedonian culinary traditions, we required a comprehensive collection of recipes in structured form. Since no such dataset existed publicly, we undertook a systematic collection and preprocessing effort to create the first structured Macedonian recipe dataset.

\subsection{Data Sources}

Our data collection focused on three prominent Macedonian cooking websites, selected for their comprehensive coverage and diverse content:

\begin{itemize}
    \item \textbf{Gotvi.mk} – A well-established platform offering extensive categorization across desserts, appetizers, salads, and main dishes, with both traditional and contemporary recipes.

    \item \textbf{MoiRecepti.mk} – A community-driven site featuring user-submitted recipes that capture home cooking traditions and regional variations.

    \item \textbf{Recepti.mk} – A structured repository providing detailed, step-by-step instructions with consistent formatting across recipe categories.
\end{itemize}

These sites were chosen to maximize the breadth of Macedonian culinary content while capturing different perspectives on recipe presentation, from professional to community-contributed content.

\subsection{Collection Process}

We implemented a systematic web scraping pipeline using Python and BeautifulSoup~\cite{richardson2007beautiful} to extract recipe data. The collection process involved iteratively scraping category and list pages to discover individual recipes, then extracting structured information from each recipe page. For each recipe, we captured:

\begin{itemize}
    \item \textbf{Title and source URL} for identification and provenance

    \item \textbf{Image URL}, when available, for potential multimodal analysis

    \item \textbf{Ingredients} as textual lists, preserving original formatting

    \item \textbf{Instructions} segmented into ordered preparation steps

    \item \textbf{Tags and categories}, when provided, for recipe classification
\end{itemize}

All collected data was stored in JSON Lines format to facilitate downstream processing and analysis, following standard practices for large-scale web data collection~\cite{bien2020recipenlg,kiddon2015mise}.

\subsection{Preprocessing and Normalization}

To ensure consistency across heterogeneous source formats, we applied several preprocessing steps while preserving the linguistic and cultural authenticity of the original content:

\begin{itemize}
    \item \textbf{Text Standardization}: Extra whitespace, line breaks, and formatting inconsistencies were normalized across ingredient lists and instructions.

    \item \textbf{Artifact Removal}: HTML remnants, numbering artifacts, and decorative symbols (such as checkbox markers in ingredient lists) were systematically removed.

    \item \textbf{Instruction Cleaning}: Sequential numbering was stripped from instruction steps (e.g., "1. Се ставаат сите состојки во сад..." → "Се ставаат сите состојки во сад..."), while preserving the natural flow of cooking directions.

\end{itemize}

At this preprocessing stage, ingredients and instructions were preserved as natural language text, deferring more sophisticated parsing of quantities, units, and semantic relationships to subsequent analysis phases.

The resulting dataset provides a substantial foundation for computational gastronomy research in the Macedonian language. With 36,237 recipes, it offers sufficient scale for pattern analysis and enables cross-cultural comparisons with established international corpora such as Recipe1M+~\cite{marin2021recipe1m+}.

The main statistics of the collected Macedonian recipe dataset are summarized in Table~\ref{tab:dataset_stats}, illustrating its scale and diversity across sources and categories.

\begin{table}[h]
\caption{Summary statistics of the Macedonian recipe dataset.}
\centering
\begin{tabular}{|l|r|}
\hline
\textbf{Metric} & \textbf{Value} \\
\hline
Total recipes collected & 36,237 \\
\hline
Unique ingredients & 26,613 \\
\hline
Average ingredients per recipe & 8 \\
\hline
Total instructions & 175,104 \\
\hline
Average instructions per recipe & 5 \\
\hline
Recipes with images & 36,196 \\
\hline
Recipes with tags & 17,373 \\
\hline
\end{tabular}
\label{tab:dataset_stats}
\end{table}

\section{Data Parsing and Preprocessing}

Following data collection, we developed a structured parsing pipeline to transform raw ingredient descriptions into machine-readable format~\cite{kiddon2015mise, haussmann2019foodkg}. This process presented unique challenges due to the variability of Macedonian culinary text, including diverse measurement systems, descriptive modifiers, and informal language patterns common in traditional recipe writing. Our primary objective was to extract structured triplets of \textbf{quantity}, \textbf{unit}, and \textbf{ingredient name} while preserving the essential semantic content of each ingredient.

\subsection{Ingredient Normalization}

Raw ingredient strings exhibited significant heterogeneity in formatting and presentation. Before structured parsing could proceed, we applied a comprehensive normalization pipeline to standardize the textual content:

\begin{itemize}
    \item \textbf{Text Standardization:} Multiple whitespace characters were collapsed to single spaces, and leading and trailing whitespace was removed to ensure consistent string processing.

    \item \textbf{Symbol Cleaning:} Decorative elements commonly found in web-scraped content, including bullets, dashes, dots, slashes, and other non-informative markers were systematically stripped while preserving meaningful punctuation.

    \item \textbf{Modifier Handling:} Optional clarifications and descriptive modifiers enclosed in parentheses or following commas were selectively removed to focus on core ingredient identification. For instance, "пилешко месо (без кожа), свежо" (chicken (without skin), fresh) was simplified to "пилешко месо" (chicken), with the understanding that modifier information could be preserved separately when needed for specific analyses.
\end{itemize}

This normalization stage was crucial for ensuring consistent downstream parsing across the diverse formatting styles encountered in our source websites.

\subsection{Quantitative Information Extraction}

Macedonian recipes employ both traditional and modern measurement systems, requiring comprehensive recognition of diverse quantity expressions~\cite{haussmann2019foodkg}. We developed specialized extraction logic for:

\begin{itemize}
    \item \textbf{Count Recognition:} Textual quantity indicators such as "една" (one), "две" (two), "пола" (half), "½" were identified and parsed at the beginning of ingredient strings.

    \item \textbf{Numerical Quantities:} Both decimal ("0.5") and fractional ("½") representations were detected and standardized to numerical values.

    \item \textbf{Unit Identification:} A comprehensive dictionary of Macedonian measurement units was compiled, including both metric units ("г" (g), "кг" (kg), "мл" (ml)) and traditional cooking measures ("шолја" (cup), "лажичка" (spoon), "кафена лажичка" (coffee spoon)).
\end{itemize}

The extraction process followed a hierarchical matching strategy, processing longer unit strings first to ensure that compound units (e.g., "кафена лажичка") were correctly identified before shorter substrings. This approach minimized parsing ambiguities and improved extraction accuracy.

\subsection{Structured Ingredient Representation}

After extracting quantitative components, the remaining text underwent additional cleaning to isolate the core ingredient name. This included removing residual parenthetical content, trailing punctuation, and any remaining decorative elements that survived the initial normalization.

The final structured representation captured each ingredient as a standardized object:

\begin{verbatim}
{
    "quantity": <numerical_value>,
    "unit": <unit_string>,
    "name": <clean_ingredient_name>
}
\end{verbatim}

For example, the raw ingredient "500г месо (телешко или свинско)" (500g meat (veal or pork)) was transformed to:

\begin{itemize}
    \item \texttt{quantity}: 500
    \item \texttt{unit}: "г" (g)
    \item \texttt{name}: "месо" (meat)
\end{itemize}

This structured approach enables downstream computational analysis while maintaining the essential quantitative and categorical information needed for recipe understanding and comparison.

\subsection{Complete Recipe Processing Pipeline}

The full parsing pipeline integrated ingredient parsing with recipe-level processing:

\begin{enumerate}
    \item \textbf{Metadata Preservation:} Recipe titles, source URLs, images, and categorical tags were cleaned and standardized while preserving all available information.

    \item \textbf{Ingredient Parsing:} Each ingredient underwent the complete normalization and extraction pipeline described above.

    \item \textbf{Data Integration:} All components were assembled into a comprehensive JSON representation suitable for computational analysis.
\end{enumerate}

This systematic approach successfully parsed the majority of ingredients in our dataset, providing a robust foundation for subsequent analysis of Macedonian culinary patterns. The resulting structured data enables both ingredient-level analysis and recipe-level comparisons that would be impossible with raw textual data.

\subsection{Challenges and Limitations}

While the parsing pipeline successfully handles the majority of ingredient strings, certain challenges remain. Ambiguous modifiers, uncommon units, or non-standard formatting occasionally lead to incomplete or imprecise extraction. For instance, ingredients with multiple optional descriptors or rare traditional measures may not be fully captured in the structured representation. These cases represent a small fraction of the dataset but highlight opportunities for future refinement. Despite these limitations, the structured data provides sufficient quality for meaningful pattern analysis, as demonstrated in  the following section.

\section{Dataset Analysis and Comparison with Recipe1M+}

Understanding the structure and characteristics of a recipe dataset is essential for both validating data quality and revealing culturally distinctive culinary patterns. To assess the Macedonian recipe collection and situate it within the broader landscape of computational gastronomy, we conducted a comparative analysis against Recipe1M+~\cite{marin2021recipe1m+}, the most widely used large-scale recipe corpus. For computational efficiency, we utilized the first 100,000 recipes from Recipe1M+ (~10\% of the full corpus)~\cite{sasanski2025aligning}. This subset captures ~70\% of unique ingredients from the complete dataset, demonstrating sufficient diversity for comparative ingredient analysis.

Our analysis addresses three key questions: (1) Which ingredients define Macedonian cuisine, and how do they compare to international collections? (2) What characteristic ingredient combinations reveal culturally specific flavor profiles and cooking practices? (3) How do statistical association measures distinguish between common pairings and culturally distinctive recipes?

We examine ingredient frequency distributions, co-occurrence patterns at the pair and triplet level, and apply Pointwise Mutual Information (PMI)~\cite{church1990word} and Lift~\cite{agrawal1993mining} to quantify association strength beyond raw frequency. This multi-faceted analysis reveals both universal culinary patterns and the distinctive features that characterize Macedonian cooking traditions.

\subsection{Ingredient Frequency Analysis}

Ingredient frequency distributions reveal both universal culinary ingredients and culturally distinctive patterns that characterize Macedonian and international recipe collections. Analysis of the top ingredients in both datasets exposes not only shared foundations but also significant divergences reflecting regional food cultures.

The Macedonian dataset is dominated by traditional baking and cooking ingredients: \textit{шеќер} (sugar, 37.0\%), \textit{сол} (salt, 35.7\%), \textit{брашно} (flour, 33.7\%), \textit{јајца} (eggs, 26.4\%), and \textit{масло} (oil, 25.0\%). Dairy products feature prominently, with \textit{млеко} (milk, 22.6\%), \textit{путер} (butter, 12.1\%), and \textit{јогурт} (yogurt, 6.5\%) all ranking highly. The presence of \textit{маргарин} (margarine, 10.4\%) in the top 15 reflects historical ingredient availability in the Balkans. Beyond common ingredients, culturally specific ones such as \textit{кашкавал} (yellow cheese, 5.1\%) and \textit{сирење} (white cheese, 5.0\%) underscore the dairy-rich character of traditional Macedonian cooking.

Recipe1M+, by contrast, exhibits patterns characteristic of Western and particularly North American cuisine. The distribution is headed by \textit{salt} (35.2\%), \textit{butter} (23.1\%), \textit{sugar} (21.2\%), \textit{olive oil} (16.5\%), and \textit{water} (15.4\%). Notable is the strong representation of baking-specific ingredients including \textit{vanilla extract} (5.9), \textit{baking powder} (7.1\%), and multiple sugar variants (\textit{brown sugar} (7.1\%), \textit{powdered sugar} (2.6\%)). The prominence of \textit{garlic cloves} (13.0\%) and \textit{olive oil} (16.5\%) reflects Mediterranean influences, while the appearance of prepared ingredients signals the dataset's inclusion of contemporary convenience cooking.

Table~\ref{tab:ingredient-frequency} presents the top 20 ingredients in both datasets with their relative frequencies. While the datasets differ substantially in size (36,237 Macedonian recipes  versus 100,000 recipes from Recipe1M+), the percentage-based rankings reveal meaningful cultural patterns.

\begin{table}[h!]
\centering
\caption{Top 20 most frequent ingredients with percentage of recipes containing each ingredient. Note: Similar ingredients appearing separately (e.g., eggs/egg, onion/onions) reflect preprocessing variations in the original datasets.}
\label{tab:ingredient-frequency}
\begin{tabular}{p{0.28\linewidth}r|p{0.28\linewidth}r}
\toprule
\multicolumn{2}{c}{MK Recipes} & \multicolumn{2}{c}{Recipe1M+} \\
\cmidrule(lr){1-2} \cmidrule(lr){3-4}
Ingredient & \% & Ingredient & \% \\
\midrule
sugar & 37.0 & salt & 35.2 \\
salt & 35.7 & butter & 23.1 \\
flour & 33.7 & sugar & 21.2 \\
eggs & 26.4 & olive oil & 16.5 \\
oil & 25.0 & water & 15.4 \\
milk & 22.6 & eggs & 14.7 \\
water & 13.7 & garlic cloves & 13.0 \\
butter & 12.1 & milk & 10.2 \\
onion & 11.8 & flour & 9.9 \\
pepper & 10.6 & onion & 9.9 \\
margarine & 10.4 & all-purpose flour & 9.5 \\
egg & 9.7 & onions & 8.8 \\
garlic & 9.4 & egg & 7.6 \\
olive oil & 7.5 & brown sugar & 7.1 \\
powdered sugar & 7.0 & baking powder & 7.1 \\
baking powder/agent & 6.8 & unsalted butter & 7.0 \\
yogurt & 6.5 & pepper & 6.7 \\
vanilla sugar & 6.3 & vegetable oil & 6.6 \\
heavy cream & 5.4 & vanilla extract & 5.9 \\
cocoa & 5.4 & salt and pepper & 5.8 \\
\bottomrule
\end{tabular}
\end{table}

Comparative analysis reveals both convergence and divergence. Universal ingredients like flour, eggs, milk, sugar, and salt dominate both collections, confirming their central role across culinary traditions. However, the datasets show clear differences in their culturally specific ingredients. Macedonian recipes prominently feature traditional Balkan dairy products such as \textit{кашкавал} (yellow cheese) and \textit{сирење} (white cheese), neither of which appears in Recipe1M+'s top 50 ingredients. The use of \textit{ванилин шеќер} (vanilla sugar) rather than liquid vanilla extract reflects regional ingredient preferences and availability. Conversely, Recipe1M+ showcases ingredients rarely found in traditional Macedonian cooking, including soy sauce, worcestershire sauce, and cream cheese, reflecting its international scope and incorporation of Asian and processed ingredients.

These frequency patterns have implications beyond descriptive statistics. The differences inform downstream computational tasks: ingredient substitution systems~\cite{shirai2021identifying} must account for regional preferences (margarine vs. butter), recipe recommendation engines~\cite{trattner2017food} should recognize culturally specific ingredients, and cross-cultural analysis must distinguish between universal ingredients and region-specific choices. Understanding which ingredients co-occur with these frequent items, explored in the following subsection, provides further insight into the structural patterns of Macedonian culinary practice.

\subsection{Ingredient Pair and Triplet Frequency}

Beyond individual ingredient frequencies, co-occurrence patterns reveal the structural building blocks of recipes and characteristic flavor combinations~\cite{ahn2011flavor}. Tables~\ref{tab:ingredient-pairs} and~\ref{tab:ingredient-triplets} present the most frequent ingredient pairs and triplets in both datasets.

\begin{table}[h!]
\centering
\caption{Top 10 most frequent ingredient pairs with percentage of recipes containing both ingredients.}
\label{tab:ingredient-pairs}
\begin{tabular}{p{0.3\linewidth}r|p{0.3\linewidth}r}
\toprule
\multicolumn{2}{c}{MK Recipes} & \multicolumn{2}{c}{Recipe1M+} \\
\cmidrule(lr){1-2} \cmidrule(lr){3-4}
Ingredient Pair & \% & Ingredient Pair & \% \\
\midrule
flour \& sugar & 17.4 & butter \& salt & 9.2 \\
flour \& eggs & 14.5 & salt \& sugar & 8.8 \\
flour \& salt & 14.2 & eggs \& salt & 7.2 \\
oil \& salt & 12.2 & salt \& water & 5.9 \\
flour \& oil & 12.0 & flour \& salt & 5.8 \\
milk \& sugar & 11.9 & butter \& sugar & 5.7 \\
flour \& milk & 11.6 & all-purpose flour \& salt & 5.5 \\
sugar \& eggs & 10.8 & butter \& eggs & 5.5 \\
salt \& eggs & 9.6 & olive oil \& salt & 5.3 \\
salt \& sugar & 9.4 & eggs \& sugar & 5.3 \\
\bottomrule
\end{tabular}
\end{table}

\begin{table}[h!]
\centering
\caption{Top 10 most frequent ingredient triplets with percentage of recipes containing all three ingredients.}
\label{tab:ingredient-triplets}
\begin{tabular}{p{0.33\linewidth}r|p{0.33\linewidth}r}
\toprule
\multicolumn{2}{c}{MK Recipes} & \multicolumn{2}{c}{Recipe1M+} \\
\cmidrule(lr){1-2} \cmidrule(lr){3-4}
Ingredient Triplet & \% & Ingredient Triplet & \% \\
\midrule
flour \& sugar \& eggs & 8.4 & eggs \& salt \& sugar & 3.2 \\
flour \& milk \& sugar & 8.0 & butter \& salt \& sugar & 3.1 \\
flour \& oil \& sugar & 6.8 & butter \& eggs \& salt & 3.0 \\
flour \& salt \& sugar & 6.6 & butter \& flour \& salt & 2.9 \\
flour \& milk \& eggs & 6.3 & flour \& salt \& sugar & 2.8 \\
flour \& oil \& salt & 6.2 & baking powder \& salt \& sugar & 2.7 \\
flour \& oil \& eggs & 5.8 & butter \& eggs \& sugar & 2.5 \\
milk \& sugar \& eggs & 5.4 & baking powder \& eggs \& salt & 2.5 \\
flour \& salt \& eggs & 5.2 & all-purpose flour \& eggs \& salt & 2.3 \\
flour \& oil \& milk & 4.5 & all-purpose flour \& baking powder \& salt & 2.2 \\
\bottomrule
\end{tabular}
\end{table}

The Macedonian dataset exhibits striking concentration around baking ingredients, with flour (\textit{брашно}) appearing in 8 of the top 10 pairs and 9 of the top 10 triplets. The most frequent pair, \textit{брашно \& шеќер} (flour \& sugar), appears in 17.4\% of the entire collection. Other dominant combinations include \textit{брашно \& јајца} (flour \& eggs, 14.5\%) and \textit{брашно \& сол} (flour \& salt, 14.2\%), reinforcing the centrality of baked goods and bread-based dishes in the dataset. The triplet \textit{брашно \& шеќер \& јајца} (flour, sugar, eggs) appears in 8.4\% of recipes, representing the foundational structure of cakes, cookies, and pastries.

Beyond baking combinations, the presence of \textit{масло \& сол} (oil \& salt, 12.2\%) and \textit{млеко \& шеќер} (milk \& sugar, 11.9\%) indicates substantial representation of savory cooking and dairy-based desserts. However, the overwhelming dominance of flour-centric combinations suggests the dataset is heavily weighted toward baking and pastry recipes, which likely reflects both the composition of the source websites and cultural preferences for home baking traditions.

Recipe1M+ displays relatively more balanced diversity compared to the Macedonian collection's extreme concentration. While baking staples remain prominent, \textit{butter \& salt} (9.2\%), \textit{salt \& sugar} (8.8\%), \textit{eggs \& salt} (7.7\%), the dataset lacks the extreme concentration around a single ingredient observed in the Macedonian collection. The appearance of \textit{olive oil \& salt} (5.3\%) signals substantial Mediterranean and savory cooking content. The presence of \textit{baking powder} in multiple top triplets reflects modern convenience baking, contrasting with the simpler flour-egg-sugar combinations typical of traditional Macedonian recipes.

A notable difference in co-occurrence density emerges from the raw counts. The Macedonian top pair appears in 17.4\% of recipes, while Recipe1M+'s top pair appears 9.2\% of its collection. This disparity reflects both dataset size differences and greater recipe diversity in the international corpus. Recipe1M+ encompasses a broader range of culinary traditions, from Asian stir-fries to European sauces to American desserts, resulting in more fragmented co-occurrence patterns. The Macedonian dataset, by contrast, shows tighter clustering around culturally specific ingredient combinations.

These frequency patterns establish which ingredient combinations are common, but they do not distinguish between universal pairings (flour and eggs appear together frequently everywhere) and culturally characteristic combinations. To identify associations that exceed chance expectation and reveal distinctive culinary patterns, we turn to statistical association measures in the following subsection.

\subsection{Ingredient Pair and Triplet Associations}

While frequency counts identify common ingredient combinations, they do not distinguish between universal pairings that appear frequently by chance and culturally characteristic associations. To identify ingredients that co-occur more often than expected, we applied two statistical association measures: Pointwise Mutual Information (PMI)~\cite{church1990word} and Lift~\cite{agrawal1993mining}.

PMI quantifies how much more (or less) likely two ingredients appear together compared to their independent probabilities:
\[
\text{PMI}(A, B) = \log_2 \frac{P(A, B)}{P(A) \cdot P(B)}
\]

Lift provides an intuitive ratio interpretation:
\[
\text{Lift}(A, B) = \frac{P(A, B)}{P(A) \cdot P(B)}
\]

A Lift value greater than 1 indicates positive association (ingredients appear together more than chance), while values less than 1 indicate negative association (ingredients appear together less than expected given their individual frequencies). These formulas extend naturally to triplets by multiplying three independent probabilities in the denominator.

To ensure statistical robustness while accounting for dataset size differences, we applied minimum occurrence thresholds of 30 recipes for the Macedonian dataset and 50 recipes for Recipe1M+. This conservative scaling maintains comparable reliability across datasets of different sizes while capturing meaningful culinary patterns~\cite{tan2004selecting}.

Tables~\ref{tab:ingredient-pmi-pairs} and~\ref{tab:ingredient-pmi-triplets} present the top 10 ingredient pairs and triplets ranked by Lift score in both datasets.

\begin{table}[h!]
\centering
\caption{Top 10 ingredient pairs by Lift score (minimum occurrence: 30 recipes for MK, 50 recipes for Recipe1M+).}
\label{tab:ingredient-pmi-pairs}
\small
\begin{tabular}{p{0.70\linewidth}rr}
\toprule
\multicolumn{3}{c}{\textbf{MK Recipes}} \\
\cmidrule(lr){1-3}
Ingredient Pair & Freq & Lift \\
\midrule
peppercorns \& bay leaf & 35 & 56.86 \\
parsnip \& celery & 43 & 45.43 \\
green pepper \& red pepper & 43 & 37.18 \\
pickles \& lean mayonnaise & 34 & 29.69 \\
wheat flour \& corn flour & 78 & 27.47 \\
coconut oil \& dates & 47 & 26.51 \\
white flour \& whole wheat flour & 45 & 26.28 \\
egg white \& egg yolk & 91 & 25.16 \\
ground walnut \& dark chocolate & 43 & 24.83 \\
parsley \& mixed dry spice & 36 & 22.38 \\
\midrule
\multicolumn{3}{c}{\textbf{Recipe1M+}} \\
\cmidrule(lr){1-3}
Ingredient Pair & Freq & Lift \\
\midrule
garlic paste \& ginger paste & 50 & 8.93 \\
cool whip topping \& jello instant vanilla pudding mix & 55 & 1.66 \\
tequila \& triple sec & 51 & 1.02 \\
pepperoni \& pizza sauce & 58 & 0.99 \\
katakuriko \& sake & 87 & 0.94 \\
dashi stock \& sake & 53 & 0.88 \\
mirin \& sake & 187 & 0.85 \\
canned pumpkin \& pumpkin pie spice & 56 & 0.78 \\
bread flour \& dry yeast & 97 & 0.65 \\
red pepper \& yellow pepper & 50 & 0.64 \\
\bottomrule
\end{tabular}
\end{table}

\begin{table}[h!]
\centering
\caption{Top 10 ingredient triplets by Lift score (min occurrence: MK=30, Recipe1M+=50).}
\label{tab:ingredient-pmi-triplets}
\begin{tabular}{p{0.70\linewidth}rr}
\toprule
\multicolumn{3}{c}{\textbf{MK Recipes}} \\
\cmidrule(lr){1-3}
Ingredient Triplet & Freq & Lift \\
\midrule
boiled eggs \& pickles \& mayonnaise & 60 & 403.32 \\
coconut oil \& honey \& oat flakes & 37 & 305.30 \\
ketchup \& oregano \& ham & 45 & 233.44 \\
egg whites \& egg yolks \& vanilla pudding & 42 & 187.90 \\
yellow cheese \& ketchup \& oregano & 86 & 160.63 \\
yellow cheese \& ketchup \& ham & 55 & 155.56 \\
white flour \& baking powder \& corn flour & 39 & 151.23 \\
seasoning \& potatoes \& red pepper & 37 & 144.40 \\
egg whites \& cornstarch \& egg yolks & 97 & 137.96 \\
boiled eggs \& mayonnaise \& sour cream & 32 & 137.20 \\
\midrule
\multicolumn{3}{c}{\textbf{Recipe1M+}} \\
\cmidrule(lr){1-3}
Ingredient Triplet & Freq & Lift \\
\midrule
lasagna noodles \& mozzarella cheese \& ricotta cheese & 65 & 2.01 \\
katakuriko \& sake \& soy sauce & 68 & 1.80 \\
mirin \& sake \& soy sauce & 149 & 1.67 \\
ginger \& mirin \& sake & 52 & 1.39 \\
dried oregano leaves \& dried thyme \& onion powder & 63 & 1.39 \\
ground coriander \& ground cumin \& ground turmeric & 56 & 0.95 \\
cayenne pepper \& dried oregano leaves \& onion powder & 64 & 0.74 \\
cayenne pepper \& dried oregano leaves \& dried thyme & 61 & 0.74 \\
dried oregano leaves \& onion powder \& paprika & 65 & 0.69 \\
dried oregano leaves \& dried thyme \& paprika & 62 & 0.69 \\
\bottomrule
\end{tabular}
\end{table}

The Macedonian dataset exhibits substantially higher Lift values (pairs: 22-57, triplets: 137-403) compared to Recipe1M+ (pairs: 0.64-8.93, triplets: 0.69-2.01). This difference reflects fundamental distinctions in dataset composition and culinary scope rather than statistical artifacts.

Macedonian top-ranked pairs reveal culturally specific cooking practices. The highest association, \textit{бибер во зрно \& ловоров лист} (peppercorns and bay leaf, Lift=56.86), represents the foundational aromatic base for traditional Balkan soups and stews. Similarly, \textit{пашканат \& целер} (parsnip and celery, Lift=45.43) appears consistently in vegetable-based dishes, while \textit{зелена пиперка \& црвена пиперка} (green and red peppers, Lift=37.18) characterizes the use of fresh produce in stuffed vegetables and salads. The appearance of specialized flour combinations (\textit{пченично брашно \& пченкарно брашно}, wheat and corn flour, Lift=27.47) reflects traditional bread-making practices mixing grain types.

Triplet associations further highlight characteristic Macedonian preparations. The combination \textit{варени јајца \& кисели краставички \& мајонез} (boiled eggs, pickles, mayonnaise, Lift=403.32) represents the standard Russian salad base ubiquitous in Balkan cuisine. Modern health-conscious baking appears in \textit{кокосово масло \& мед \& овесни снегулки} (coconut oil, honey, oat flakes, Lift=305.30), while pizza and sandwich preparations manifest in \textit{кечап \& оригано \& шунка} (ketchup, oregano, ham, Lift=233.44).

Recipe1M+ exhibits more moderate association scores, reflecting its diverse international scope. The strongest pair association, \textit{garlic paste \& ginger paste} (Lift=8.93), identifies the aromatic foundation of Indian and Southeast Asian cooking. However, many frequent combinations show Lift values below 1, indicating negative associations. For instance, \textit{mirin \& sake} (Lift=0.85) and \textit{bread flour \& dry yeast} (Lift=0.65) appear together in many recipes but less often than their individual frequencies would predict. This occurs because these ingredients are common across diverse recipe types: sake appears in numerous Japanese dishes beyond those requiring mirin, while bread flour is used in many baking contexts that do not require yeast.

Triplet patterns in Recipe1M+ similarly reflect broad culinary diversity. The highest-ranked triplet, \textit{lasagna noodles \& mozzarella cheese \& ricotta cheese} (Lift=2.01), identifies a classic Italian preparation. Japanese cooking bases appear in multiple entries (\textit{katakuriko \& sake \& soy sauce}, \textit{mirin \& sake \& soy sauce}), while spice blends represent various traditions (\textit{dried oregano leaves \& dried thyme \& onion powder}, \textit{ground coriander \& ground cumin \& ground turmeric}).

The contrast between datasets reveals how association measures capture culinary coherence. Macedonian recipes, drawn from a relatively homogeneous culinary tradition, exhibit tight ingredient clustering around culturally specific practices. Recipe1M+'s international diversity disperses ingredient usage across multiple culinary contexts, resulting in lower association scores even for recognized combinations. This difference has implications for computational applications: ingredient substitution systems must account for whether replacements maintain cultural authenticity within a tradition (high-association context) or cross-cultural adaptability (low-association, diverse context). Similarly, recipe generation models trained on culturally specific datasets will produce more cohesive traditional dishes, while models trained on diverse corpora generate more fusion-oriented outputs.

\section{Discussion and Limitations}

This work presents the first structured dataset of Macedonian recipes and demonstrates how computational methods can reveal characteristic patterns in regional culinary traditions. The comparative analysis with Recipe1M+ illustrates both the distinctive features of Macedonian cuisine and the broader methodological challenges of representing underrepresented food cultures in computational gastronomy~\cite{min2019survey}.

\subsection{Cultural Coherence vs. International Diversity}

The most striking finding from our analysis is the fundamental difference in ingredient association patterns between the two datasets. Macedonian recipes exhibit substantially higher Lift values (pairs: 22-57, triplets: 137-403) compared to Recipe1M+ (pairs: 0.64-8.93, triplets: 0.69-2.01), reflecting tightly clustered ingredient usage around culturally specific practices. This is not a statistical artifact but a meaningful signal of culinary coherence.

Several factors contribute to this pattern. First, the Macedonian dataset draws from a relatively homogeneous culinary tradition with shared historical roots, ingredient availability, and cooking techniques. Ingredients like peppercorns and bay leaves consistently appear together because they form the aromatic foundation of traditional Balkan soups and stews, a practice reinforced across generations. Second, the source websites predominantly feature home cooking and traditional recipes rather than fusion cuisine or international adaptations, further concentrating ingredient usage patterns.

Recipe1M+, by contrast, aggregates recipes from diverse global traditions, including Italian pasta, Thai curries, American baking, and Indian spice blends, resulting in ingredients appearing across multiple unrelated culinary contexts. Even classic combinations like \textit{mirin \& sake} show Lift values below 1.0 because these ingredients appear frequently in different Japanese recipe types that don't overlap. This dispersion reflects genuine international diversity rather than weak culinary associations.

The comparison reveals a fundamental tension in computational gastronomy: datasets capturing specific traditions enable deep cultural understanding but limited generalization, while internationally diverse corpora support broad applications but obscure regional specificity. Both perspectives are valuable for different research questions: recipe recommendation systems benefit from diverse training data, while cultural preservation and traditional recipe generation require focused, coherent collections.

\subsection{Methodological Considerations and Threshold Selection}

Our analysis applied minimum occurrence thresholds of 30 recipes for the Macedonian dataset and 50 recipes for Recipe1M+ to balance statistical robustness with pattern discovery~\cite{tan2004selecting}. These thresholds represent roughly 0.08\% and 0.005\% of each dataset respectively, capturing combinations that are neither universal ingredients nor extreme outliers.

The choice of these specific thresholds involved trade-offs. Lower thresholds (e.g., min=5) produced extreme Lift values exceeding 50 million for Macedonian triplets, mathematically correct but reflecting data sparsity rather than culinary insight. Higher thresholds (e.g., min=100) would exclude culturally meaningful but moderately frequent combinations. Our selected thresholds successfully identified interpretable patterns: the Russian salad base (eggs, pickles, mayonnaise) and soup aromatics (peppercorns, bay leaves) in Macedonian recipes, and lasagna ingredients and Asian cooking bases in Recipe1M+.

Many Recipe1M+ pairs exhibit negative PMI values and Lift scores below 1.0, indicating that these ingredients co-occur less frequently than their individual frequencies predict, not because they are incompatible, but because they are versatile. Common ingredients like \textit{warm water} appear in thousands of recipe contexts, so their co-occurrence with specific ingredients like \textit{active dry yeast} is diluted despite being a classic baking combination. This demonstrates how association measures capture different information than raw frequency: \textit{active dry yeast \& warm water} is frequent but not distinctive, while \textit{garlic paste \& ginger paste} is both frequent and characteristic of specific cuisines.

\subsection{Dataset Composition and Coverage Limitations}

The Macedonian dataset, while substantial at 36,237 recipes, reflects the coverage and biases of its three source websites. Several limitations arise from this composition:

\textbf{Thematic concentration:} The dominance of flour-containing pairs (8 of top 10) and baking-related triplets suggests overrepresentation of desserts and baked goods relative to other meal categories. This likely reflects both the source websites' content focus and cultural preferences for home baking, but may not fully represent the breadth of Macedonian cuisine including soups, grilled meats, and preserved vegetables.

\textbf{Modern vs. traditional balance:} While the dataset captures traditional recipes well, contemporary fusion cooking, restaurant preparations, and adaptations of international dishes are underrepresented. This limits insights into how Macedonian cuisine is evolving in response to globalization and changing ingredient availability.

\textbf{Regional variation:} While a subset of recipes includes regional or category tags, this metadata is incomplete and inconsistently structured across sources, limiting systematic analysis of geographic patterns within Macedonia or comparative studies with neighboring Balkan cuisines.

\textbf{Temporal coverage:} The dataset represents a snapshot of currently available online recipes, with no historical dimension to track culinary evolution over time.

These limitations are common to web-scraped recipe datasets but particularly affect smaller collections focused on underrepresented languages. Expanding beyond the current three sources to include food blogs, digitized cookbooks, and community submissions would improve representativeness.

\subsection{Technical Parsing Challenges}

The ingredient parsing pipeline successfully structured the majority of recipes but faces ongoing challenges:

\textbf{Morphological variation:} Macedonian's inflectional morphology produces multiple forms of the same ingredient (\textit{јајца} vs. \textit{јајце} for eggs, \textit{кромид} vs. \textit{лук} for onion/garlic). While we addressed common variations, systematic lemmatization would improve frequency counts and association measures.

\textbf{Quantity ambiguity:} Expressions like "1-2 cups," "a pinch of," or "по вкус" (to taste) resist structured extraction. We preserved these as textual annotations but they complicate quantitative analysis of ingredient proportions.

\textbf{Compound ingredients:} Multi-word ingredient names (\textit{маслиново масло за прелив} (olive oil for dressing)) sometimes confuse ingredient identity with preparation context, artificially fragmenting the ingredient space.

These issues are not unique to Macedonian. Recipe1M+ exhibits similar challenges with ingredient name variants (\textit{onion} vs. \textit{onions}, \textit{all-purpose flour} vs. \textit{flour}), but they are harder to address in lower-resource languages lacking comprehensive culinary lexicons or trained NLP models.

\subsection{Implications for Computational Gastronomy}

Despite these limitations, the dataset and analysis contribute several insights relevant to broader computational gastronomy research:

\textbf{Cross-cultural recipe understanding:} Association measures reveal that ingredient relationships are culturally situated. A recipe generation model trained solely on Recipe1M+ would struggle to produce authentic Macedonian dishes~\cite{salvador2019inverse} because it lacks the tight ingredient clustering that defines traditional preparations. Conversely, models trained on culturally specific datasets produce more coherent traditional recipes but cannot easily generalize to fusion or international contexts.

\textbf{Ingredient substitution systems:} Effective substitution requires understanding whether ingredients are universally interchangeable or culturally essential~\cite{shirai2021identifying, majumder2019generating}. High-Lift pairs in Macedonian recipes (e.g., wheat and corn flour combinations) represent practiced traditions that should be preserved, while lower-association ingredients offer more flexibility for adaptation.

\textbf{Nutritional and health analysis:} The dataset enables population-specific dietary studies that international corpora cannot support. Understanding actual ingredient usage patterns in traditional Macedonian cooking, rather than idealized or restaurant-style preparations, provides a more accurate foundation for nutritional recommendations and public health interventions~\cite{min2019survey}.

\textbf{Low-resource language modeling:} This work demonstrates feasible methods for constructing structured food datasets in languages underrepresented in NLP research, contributing to efforts to diversify training data beyond English and other major languages.

\subsection{Future Directions}

Several concrete extensions would strengthen this research:

\textbf{Expanded data collection:} Incorporating additional Macedonian recipe sources (blogs, social media, digitized cookbooks) and enriching metadata (regional tags, meal types, seasonal indicators) would improve coverage and enable finer-grained analyses.

\textbf{Enhanced linguistic processing:} Developing or adapting morphological analyzers, building bilingual culinary lexicons, and applying embedding-based semantic matching would address current normalization limitations.

\textbf{Temporal and evolutionary analysis:} Collecting historical recipes or timestamped online content would enable tracking how ingredient usage patterns change with modernization, globalization, and ingredient availability.

\textbf{Cross-regional comparison:} Extending the methodology to neighboring Balkan countries (Serbia, Bulgaria, Greece) would reveal shared culinary patterns and distinctive regional variations within a related cultural sphere.

The observed ingredient patterns in this dataset are not merely statistical regularities; they encode authentic culinary knowledge practiced in Macedonian kitchens. Each strong association represents a tested flavor combination, and each frequent triplet a reliable recipe structure. As the dataset expands and analytical methods mature, these signals will become increasingly valuable for preserving culinary heritage, supporting recipe innovation, and advancing the computational study of food cultures beyond the limited set of languages and cuisines currently represented in research datasets.

\section{Conclusion}

We presented the first structured dataset of Macedonian recipes, comprising 36,237 recipes with parsed ingredients and cooking instructions. Through systematic web scraping, preprocessing, and statistical analysis, we demonstrated that computational methods can effectively capture and quantify culinary patterns in underrepresented languages and food cultures.

Our comparative analysis with Recipe1M+ revealed fundamental differences between culturally focused and internationally diverse recipe collections. Macedonian recipes exhibit substantially higher ingredient association scores (Lift: 22-403) than Recipe1M+ (Lift: 0.64-8.93), reflecting tight clustering around traditional cooking practices rather than statistical artifacts. This contrast illustrates how dataset composition shapes observable patterns: homogeneous culinary traditions produce strong, interpretable associations, while international diversity disperses ingredient usage across multiple unrelated contexts.

The dataset and methodology enable several contributions to computational gastronomy. First, it extends the geographical and linguistic coverage of structured recipe data beyond the English-dominated landscape of existing resources. Second, the ingredient co-occurrence patterns provide empirical grounding for understanding Balkan culinary structure, supporting applications from recipe recommendation to cultural preservation. Third, the parsing pipeline and association analysis framework offer replicable methods for constructing similar datasets in other underrepresented languages.

This work demonstrates that even modestly-sized datasets from specific culinary traditions can reveal authentic patterns that large-scale international corpora obscure. As computational gastronomy expands beyond its current narrow focus on major languages and cuisines, such regionally focused resources will become essential for building truly inclusive and culturally aware food computing systems.

\section{Code Availability}
The code for dataset generation, parsing, and analysis is available at \url{https://gitlab.finki.ukim.mk/ds4food/MKRecipes}.
\\

\textbf{Acknowledgement:} This work was supported by the Ministry of Education and Science of the Republic of North Macedonia through the project "Utilizing AI and National Large Language Models to Advance Macedonian Language Capabilities".

\bibliography{cas-ref}

\end{document}